\documentclass[10pt,twocolumn,letterpaper]{article}

\usepackage{iccv}              % To produce the CAMERA-READY version
\usepackage{multirow}
\usepackage{tabularx}
\usepackage{amssymb}

\usepackage{float} % Yoonki
\usepackage{caption} 
\usepackage{capt-of}
\usepackage{comment}
\definecolor{iccvblue}{rgb}{0.21,0.49,0.74}
\usepackage[pagebackref,breaklinks,colorlinks,allcolors=iccvblue]{hyperref}

\def\eg{\emph{e.g}\onedot, }

\def\etal{\emph{et al}\onedot}

\def\paperID{*****} % *** Enter the Paper ID here
\def\confName{ICCV}
\def\confYear{2025}

\title{Pinpointing Trigger Moment for Grounded Video QA: \\ Enhancing Spatio-temporal Grounding in Multimodal Large Language Models}

\author{
Jinhwan Seo$^1$\quad
Yoonki Cho$^1$\quad
Junhyug Noh$^2$\quad
Sung-Eui Yoon$^1$\\[0.25em]
$^1$KAIST\quad
$^2$Ewha Womans University\\
{\tt\small \{jinhwan.seo, yoonki, sungeui\}@kaist.ac.kr}\quad
{\tt\small junhyug@ewha.ac.kr}
}

\begin{document}
\maketitle

\begin{abstract}
In this technical report, we introduce a framework to address Grounded Video Question Answering (GVQA) task for the ICCV 2025 Perception Test Challenge. The GVQA task demands robust multimodal models capable of complex reasoning over video content, grounding the resulting answers visually, and tracking the referenced objects temporally. To achieve this capability, our proposed approach decomposes the GVQA task into a three-stage pipeline: (1) Video Reasoning \& QA, (2) Spatio-temporal Grounding and (3) Tracking. Our key contribution is the introduction of a trigger moment, derived from our proposed CORTEX prompt, which pinpoints the single most visible frame of a target object to serve as a robust anchor for grounding and tracking. To this end, we achieve the HOTA score of 0.4968, which marks a significant improvement over the previous year's winning score of 0.2704 on GVQA task.
\end{abstract}

\section{Introduction}
\label{sec:intro}
The rapid development of Multimodal Large Language Model (MLLM) and vision-language model (VLM) has resulted in significant advancements in multimodal reasoning about complex visual content. 
Despite remarkable progress in achieving high QA performance, researchers~\cite{xiao2024can,patraucean2023perception,liu2024timecraft,chen2025cross,xu2024exploring} reveal that existing VideoQA models may rely heavily on language shortcuts instead of learning from the necessary causal visual content. This issue highlighted the need for a task that compels models to justify their answers with visual evidence. In response, Grounded Video Question Answering (GVQA) was introduced. GVQA represents a critical task in advancing foundational multimodal models, requiring not only to reason about visual content but also to associate conclusions with corresponding visual evidence.

Patraucean~\etal~\cite{patraucean2023perception} introduced a diagnostic benchmark, \textbf{Perception Test}, to recognize the critical need for diagnostic evaluation of multimodal models, particularly to measure generalization capabilities. This benchmark is explicitly designed to comprehensively assess multimodal perception and reasoning skills, spanning various modalities (video, audio, and text). The perception test includes dedicated GVQA task, which is especially designed to connect high-level semantic understanding and low-level perception skills, requiring answers to be provided as object tracks associated with the query.

In this report, we propose a three-stage pipeline centered on our main contribution: \textit{trigger moment}. This technique entails leveraging the chain-of-thought reasoning of an MLLM to identify a single, key frame in which the target object is visible. The introduction of trigger moment as spatio-temporal anchor leads to a significant enhancement in grounding performance.
%The GVQA differs from conventional Video Question Answering (VideoQA) because the output is not merely a textual answer but requires bounding boxes tracking targeted objects throughout the video. This necessity mandates the fusion of high-level semantic reasoning with low-level visual perception tasks such as detection and tracking.

%-------------------------------------------------------------------------

\begin{figure*}[ht!]
\centering
\includegraphics[trim=0cm 0cm 0cm 0cm,clip,width=\textwidth]{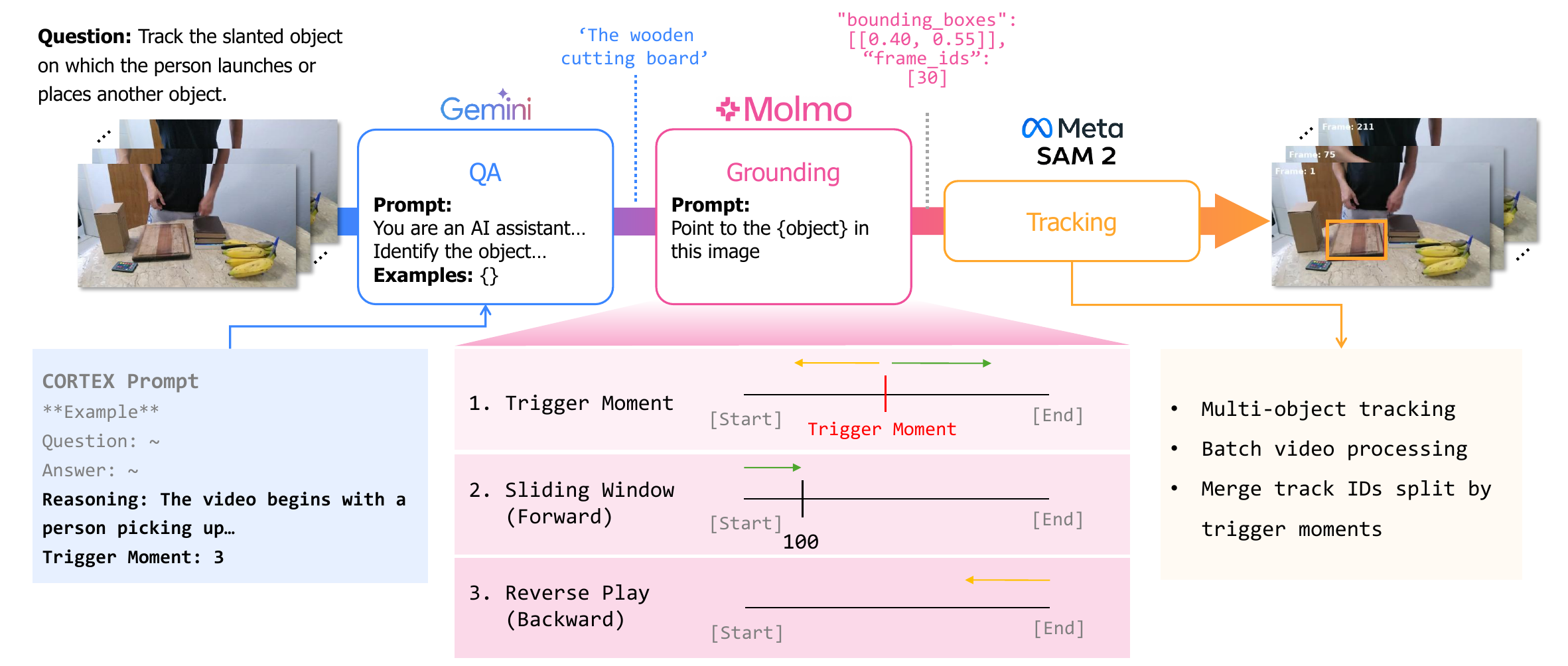}
\caption{
An overview of the proposed approach. The QA stage uses a CORTEX prompt with Gemini~\cite{comanici2025gemini} to produce fine-grained reasoning and identify a \textit{trigger moment}, an optimal frame for localization. In the Grounding stage, Molmo~\cite{deitke2025molmo} localizes the object within this trigger moment to generate an initial bounding box. If grounding fails at trigger moment, the model employs fallback strategies, such as a sliding window and reverse play search, to find an alternative frame to initialize the grounding. Finally, SAM2~\cite{ravi2024sam} produces the full trajectory by initiating from grounded point and applying bidirectional tracking.
}
\label{fig:arch}
\end{figure*}

\section{Method}
\label{sec:method}
Our approach to the Grounded VideoQA (GVQA) challenge is structured as a three-stage pipeline designed to causal reasoning about tracking objects based on a natural language query. As illustrated in~\cref{fig:arch}, three-stages consist of: (1) Video Reasoning \& QA, which identifies the target objects described in the question; (2) Grounding, which localizes the identified objects within a specific video frame; and (3) Tracking, which propagates this initial point throughout the video to complete a trajectory.

%-------------------------------------------------------------------------
\subsection{Video Reasoning \& QA}
The initial stage of our pipeline is responsible for comprehending the given question and identifying the target objects within the context of the video. For the high-level reasoning task, we utilize Gemini 2.5 Pro~\cite{comanici2025gemini}, a powerful multimodal model with advanced reasoning capabilities. We provide the model with video and the textual question. To achieve this, we employ a few-shot prompting strategy that we term CORTEX (Chain-of-Reasoning for Trigger-moment Extraction). The CORTEX prompt is designed to guide the model not only to output a concise, descriptive text string that uniquely recognizes the target object but also to elicit the model's underlying reasoning process. Crucially, the prompt instructs the model to pinpoint a \textit{trigger moment}: a single, 0-indexed frame where the identified object is most clearly and fully visible, considering factors like focus, lighting, and occlusion.
For example, for the question \textit{``Track the slanted object on which the person launches or places another object,''} the model is expected to output a specific description like \textit{``The wooden cutting board.''} This textual answer serves as the input for the subsequent grounding stage.

\noindent\textbf{OCR Cases.} The grounding model often struggled to localize small text objects, such as individual letters in words like \textit{`h', `o', `m', `e'} for questions like \textit{``Track the letters that the person interacts with.''} Therefore, for OCR cases, we added a descriptive form of output to the prompt that could be distinguished instead of spelling.
\begin{figure}[ht]
\vspace{-0.3cm}
\includegraphics[trim=0cm 0cm 0cm 0cm,clip,width=\columnwidth]{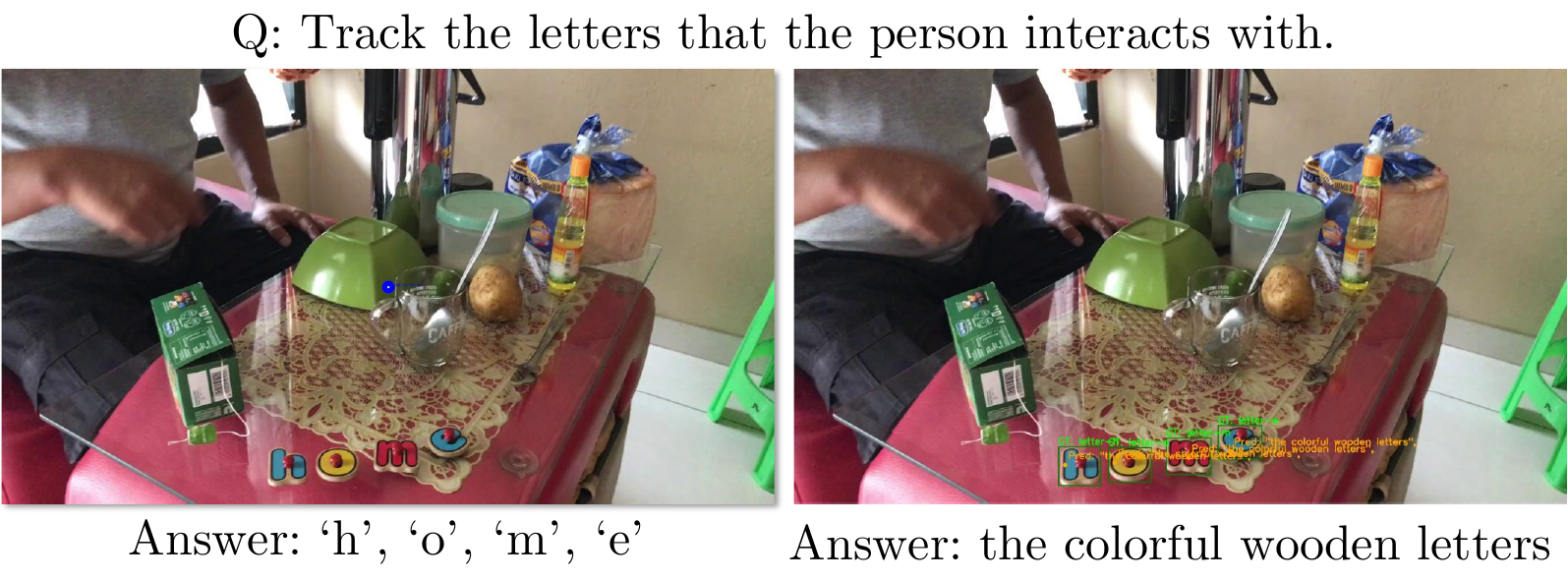}
\vspace{-0.6cm}
\caption{
OCR failure cases.
}
\label{fig:limit}
\vspace{-0.3cm}
\end{figure}

\subsection{Spatio-temporal Grounding}
Once the target object is identified as a text description, the grounding stage localizes it within a single video frame, providing an initial spatio-temporal anchor. For this task, we use Molmo~\cite{deitke2025molmo}, an open-weight vision-language model (VLM) adept at fine-grained visual understanding and localization. The Molmo model is prompted with a single frame from the video and a ``point-to-box'' instruction, such as \textit{``Point to the \{object name\} in this image,''} where the object description is the text generated by the Gemini model. Molmo then outputs the normalized coordinates for a bounding box that encloses the target object in that specific frame. In our proposed method, trigger moment, we first attempt to ground the object at trigger moment. If initial attempt fails, we employ a fallback mechanism, reverting to the sliding window and reverse play strategies to find a initial frame.

\noindent\textbf{Naive Sliding Window.}
A key challenge in GVQA is handling questions that refer to events based on temporal grounding. Our first approach for spatio-temporal grounding was a naive sliding window. This method involved attempting to ground the object using Molmo~\cite{deitke2025molmo} starting from the first frame of the video. If it failed, it would skip forward by a 30-frame interval and try again, repeating this search until the object was successfully grounded.

\noindent\textbf{Reverse Play.}
However, we observed that the sliding window method often led to false positives. It was easily deceived in cases with decoy objects, such as \textit{``Track the letters put by the person on table''} case, where a decoy appears first. We also identified issues in cases where target objects, like letters, were overlapping in the initial frames. To resolve these issues, we adopted a reverse play strategy. If the target object is not found within the first 100 frames, we start searching for the initial box from the end of the video, moving backward.

\noindent\textbf{Trigger Moment.}
Motivated by Chain-of-Thought (CoT), we refined our approach to explicitly leverage the reasoning process of the QA model. We prompted Gemini not only to provide the final answer but also to output the reasoning that led to it. We then used this textual reasoning to select a Trigger Moment. We found that this trigger moment is highly suitable as an initial frame because it effectively pinpoints a frame where the target object is present and, crucially, identifies the key frame the model referenced to derive its answer. Once trigger moment is identified, the tracking can be performed bidirectionally from trigger moment frame. This approach is optimized for the challenge's format as it maintains a association ID for specific object originating from a single video frame. This significant advantage over alternative methods, such as merging multiple tracklets which can lose the original association ID or performing computationally expensive grounding on every frame.
\begin{figure}[ht]
\vspace{-0.3cm}
\includegraphics[trim=0cm 0cm 0cm 0cm,clip,width=\columnwidth]{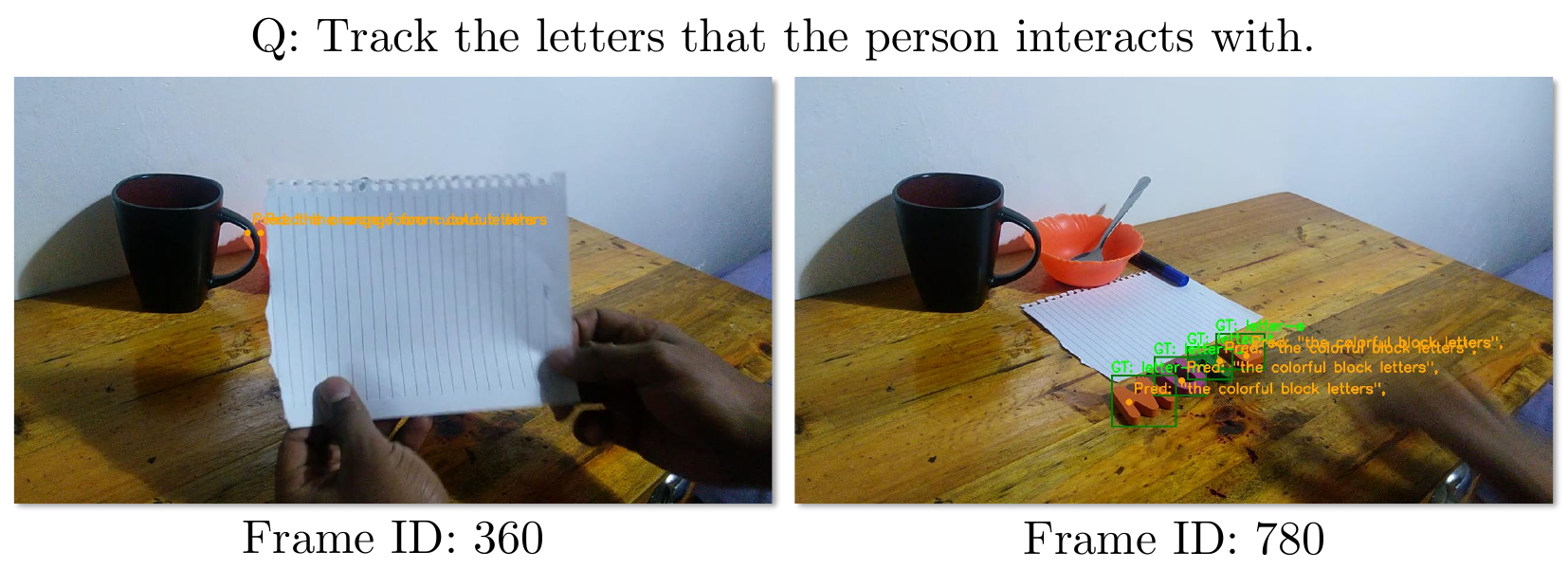}
\vspace{-0.6cm}
\caption{
Comparision of temporal grounding (Left: Naive Sliding Window, Right: Trigger Moment).
}
\label{fig:temporal}
\vspace{-0.3cm}
\end{figure}

\subsection{Tracking}
The final stage of our pipeline takes the initial bounding box from the grounding model and generates a full trajectory for the object throughout the video. To do this, we employ SAM2~\cite{ravi2024sam}, a state-of-the-art model for video segmentation and tracking. Since the output from grounding model is a point coordinate, we provide this to SAM2 using a point-to-box prompt. SAM2 then propagates an initial bounding box from this point and proceed the localization across the entire video sequence to produce a tracklet, which is a sequence of bounding boxes corresponding to the object's position in each frame. We chose SAM2 for its open-weights availability and its robust performance in multi-object tracking scenarios.

\section{Experiment}
\label{sec:experiment}

\noindent\textbf{Dataset.}
The Grounded VideoQA dataset for the Perception Test challenge consists of a training set with $586$ videos and $1{,}859$ questions, a validation set with $1{,}545$ videos and $3{,}051$ questions, and a test set with $932$ videos and $1{,}859$ questions. On average, videos in the training and validation sets contained about $708$ frames, while the test set videos averaged $611$ frames.

\noindent\textbf{Evaluation Metric.}
The performance for GVQA challenge was evaluated using the HOTA~\cite{luiten2021hota} metric. To achieve a high HOTA score, a model must not only accurately detect objects but also consistently maintain identities throughout the video sequence.

\noindent\textbf{Implementation Details.}
Our method is training-free; we use pre-trained models and evaluate the performance in zero-shot manner. For the proposed approach, we utilized Gemini 2.5 Pro~\cite{comanici2025gemini} for the QA stage, providing it with the video downsampled to $3$ fps, the corresponding question, and a few-shot prompt. For the grounding stage, we employed Molmo-7B~\cite{deitke2025molmo}. To implement the naive sliding window, we detected failure signals from the model. We also considered it a failure when more than seven objects were detected in a single frame, prompting a skip to the next frame. The frame threshold for initiating the reverse play strategy was set to $100$. In the final tracking stage, we used SAM2~\cite{ravi2024sam} with its \texttt{hiera-large} pre-trained weights and processed the video at $10$ fps to generate the trajectories.

\begin{table}[t!]
    \centering
    \small
    \setlength{\tabcolsep}{6pt}
    \caption{Ablation study on \textit{validation} set. As a baseline method, we utilized Gemini 2.5 Flash~\cite{comanici2025gemini} + Molmo-7B~\cite{deitke2025molmo} + SAM2~\cite{ravi2024sam}. On top of this baseline, we evaluated the effectiveness of our proposed method for spatio-temporal grounding.}
    \vspace{-0.3cm}
    \resizebox{\columnwidth}{!}{
    \begin{tabular}{ccc|c}
        \toprule
        Sliding Window & Reverse & Trigger Moment & HOTA ($\uparrow$) \\
        \midrule
        \checkmark & &  &0.3687 \\
        \checkmark & \checkmark & & 0.4011 \\
        \checkmark & \checkmark & \checkmark & 0.4380 \\
        \bottomrule
    \end{tabular}} 
    \label{tab:ablation_study}
\end{table}

\begin{table}[t!]
    \centering
    \small
    \setlength{\tabcolsep}{10pt}
    \caption{Comparison on the \textit{test} set and leaderboard ranking.}
    \vspace{-0.3cm}
    \begin{tabular}{c l c}
        \toprule
        Rank & Method & HOTA ($\uparrow$) \\
        \midrule
         & Baseline (MDETR-static)~\cite{patraucean2023perception} & 0.0574 \\
         & 2024 Winner~\cite{heyward2024perception} & 0.2704 \\
        \midrule
        \bf 1 & \bf SGVR@KAIST (Ours) & \bf 0.4968 \\
        2 & TutuAI            & 0.4304 \\
        3 & NJUST-KMG         & 0.4002 \\
        4 & lababa            & 0.3304 \\
        5 & aolaxing          & 0.2475 \\
        \bottomrule
    \end{tabular}
    \label{tab:sota}
\end{table}

\noindent\textbf{Results.}
We conducted experiments with various combinations of models and strategies, with the HOTA scores reported on the validation set. Please note that the results reported on validation set in~\cref{tab:ablation_study} are from experiments for ablation study during the competition, not from the final prompt configuration. Our baseline setup, using Gemini 2.5 Flash, Molmo-7B, and multi-object tracking configuration, the score using a sliding window for grounding achieved a HOTA score of $0.3687$. To address the issue of decoy objects appearing in the early frames, we implemented the Reverse Play strategy, which improved the score to $0.4011$. Our final proposed method, leveraging a Trigger Moment derived from the QA model's reasoning, demonstrated the best performance, achieving a HOTA score of $0.4380$ on the validation set as shown in~\cref{tab:ablation_study}.
On the test set, our final model using Trigger Moment achieved a remarkable HOTA score of $0.4968$. This result significantly surpasses previous state-of-the-art performances, including the baseline score of $0.0574$ from MDETR-static and the 2024 winner's score of $0.2704$. These results confirm the effectiveness of our three-stage pipeline and highlight the superiority of the Trigger Moment grounding strategy for Grounded VideoQA task.

\section{Discussion}
Our method demonstrates significant performance, but there are areas for further exploration. The proposed approach, centered on a single trigger moment, is effective for singular events. However, for questions involving multiple events (\eg \textit{``Track the objects shown to the camera more than once''}), a potential future direction is multi-stage process that handles multiple trigger moments to form a comprehensive answer. Additionally, we noted that questions requiring complex physical reasoning (\eg \textit{``From the containers that the person pours into, which one is the tallest?''}) occasionally challenged the QA model's capabilities, presenting an opportunity for further refinement in nuanced, comparative understanding.

\begin{figure}[ht]
\vspace{-0.3cm}
\includegraphics[trim=0cm 0cm 0cm 0cm,clip,width=\columnwidth]{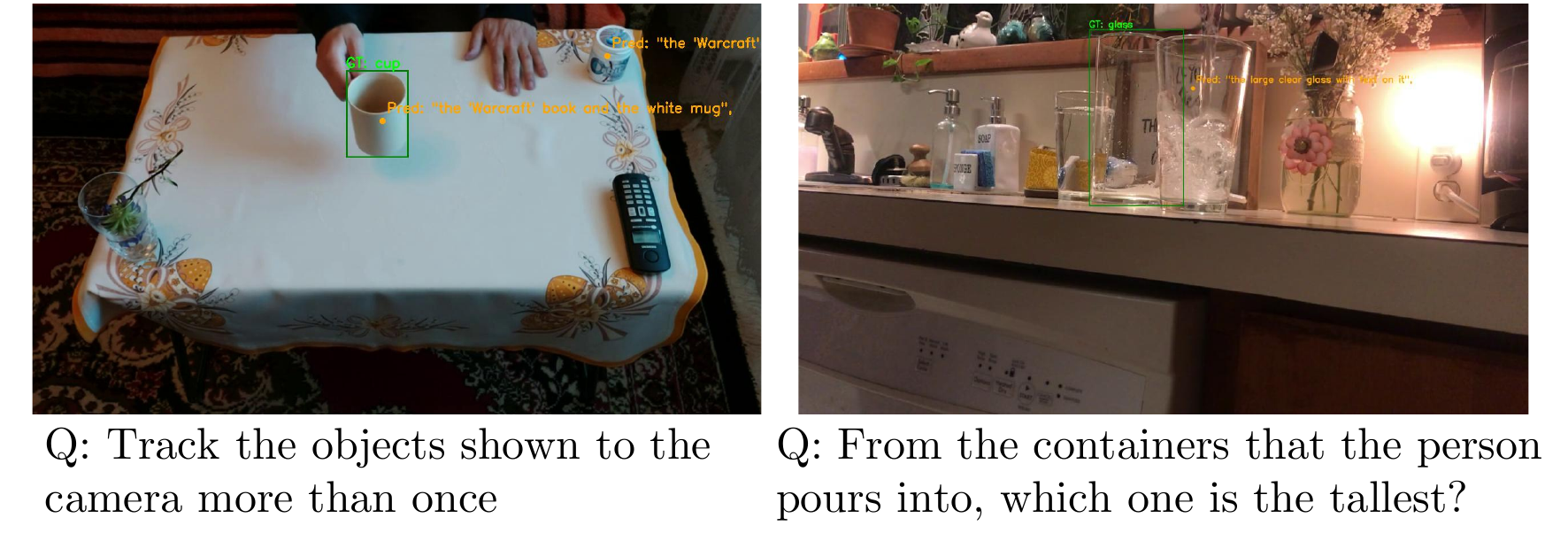}
\vspace{-0.6cm}
\caption{
Failure cases.
}
\label{fig:limit}
\end{figure}

\smallskip
\noindent\textbf{Acknowledgements}.
This work was supported by the Institute of Information \& communications Technology Planning \& Evaluation (IITP) grant funded by the Korea government (MSIT) (No. RS-2025-25443318, Physically-grounded Intelligence: A Dual Competency Approach to Embodied AGI through Constructing and Reasoning in the Real World; and No. RS-2023-00237965, Recognition, Action and Interaction Algorithms for Open-world Robot Service) and the National Research Foundation of Korea (NRF) grant funded by the Korea government (MSIT) (No. RS-2023-00208506).
{
    \small
    \bibliographystyle{ieeenat_fullname}
    \bibliography{main}

\begin{thebibliography}{10}
\providecommand{\natexlab}[1]{#1}
\providecommand{\url}[1]{\texttt{#1}}
\expandafter\ifx\csname urlstyle\endcsname\relax
  \providecommand{\doi}[1]{doi: #1}\else
  \providecommand{\doi}{doi: \begingroup \urlstyle{rm}\Url}\fi

\bibitem[Chen et~al.(2025)Chen, Liu, Chen, Su, Zheng, and Lin]{chen2025cross}
Weixing Chen, Yang Liu, Binglin Chen, Jiandong Su, Yongsen Zheng, and Liang Lin.
\newblock Cross-modal causal relation alignment for video question grounding.
\newblock In \emph{Proceedings of the Computer Vision and Pattern Recognition Conference}, pages 24087--24096, 2025.

\bibitem[Comanici et~al.(2025)Comanici, Bieber, Schaekermann, Pasupat, Sachdeva, Dhillon, Blistein, Ram, Zhang, Rosen, et~al.]{comanici2025gemini}
Gheorghe Comanici, Eric Bieber, Mike Schaekermann, Ice Pasupat, Noveen Sachdeva, Inderjit Dhillon, Marcel Blistein, Ori Ram, Dan Zhang, Evan Rosen, et~al.
\newblock Gemini 2.5: Pushing the frontier with advanced reasoning, multimodality, long context, and next generation agentic capabilities.
\newblock \emph{arXiv preprint arXiv:2507.06261}, 2025.

\bibitem[Deitke et~al.(2025)Deitke, Clark, Lee, Tripathi, Yang, Park, Salehi, Muennighoff, Lo, Soldaini, et~al.]{deitke2025molmo}
Matt Deitke, Christopher Clark, Sangho Lee, Rohun Tripathi, Yue Yang, Jae~Sung Park, Mohammadreza Salehi, Niklas Muennighoff, Kyle Lo, Luca Soldaini, et~al.
\newblock Molmo and pixmo: Open weights and open data for state-of-the-art vision-language models.
\newblock In \emph{Proceedings of the Computer Vision and Pattern Recognition Conference}, pages 91--104, 2025.

\bibitem[Heyward et~al.(2024)Heyward, Carreira, Damen, Zisserman, and P{\u{a}}tr{\u{a}}ucean]{heyward2024perception}
Joseph Heyward, Jo{\~a}o Carreira, Dima Damen, Andrew Zisserman, and Viorica P{\u{a}}tr{\u{a}}ucean.
\newblock Perception test 2024: Challenge summary and a novel hour-long videoqa benchmark.
\newblock \emph{arXiv preprint arXiv:2411.19941}, 2024.

\bibitem[Liu et~al.(2024)Liu, Ma, Zhong, Zhang, and Lin]{liu2024timecraft}
Huabin Liu, Xiao Ma, Cheng Zhong, Yang Zhang, and Weiyao Lin.
\newblock Timecraft: Navigate weakly-supervised temporal grounded video question answering via bi-directional reasoning.
\newblock In \emph{European Conference on Computer Vision}, pages 92--107. Springer, 2024.

\bibitem[Luiten et~al.(2021)Luiten, Osep, Dendorfer, Torr, Geiger, Leal-Taix{\'e}, and Leibe]{luiten2021hota}
Jonathon Luiten, Aljosa Osep, Patrick Dendorfer, Philip Torr, Andreas Geiger, Laura Leal-Taix{\'e}, and Bastian Leibe.
\newblock Hota: A higher order metric for evaluating multi-object tracking.
\newblock \emph{International journal of computer vision}, 129\penalty0 (2):\penalty0 548--578, 2021.

\bibitem[Patraucean et~al.(2023)Patraucean, Smaira, Gupta, Recasens, Markeeva, Banarse, Koppula, Malinowski, Yang, Doersch, et~al.]{patraucean2023perception}
Viorica Patraucean, Lucas Smaira, Ankush Gupta, Adria Recasens, Larisa Markeeva, Dylan Banarse, Skanda Koppula, Mateusz Malinowski, Yi Yang, Carl Doersch, et~al.
\newblock Perception test: A diagnostic benchmark for multimodal video models.
\newblock \emph{Advances in Neural Information Processing Systems}, 36:\penalty0 42748--42761, 2023.

\bibitem[Ravi et~al.(2024)Ravi, Gabeur, Hu, Hu, Ryali, Ma, Khedr, R{\"a}dle, Rolland, Gustafson, et~al.]{ravi2024sam}
Nikhila Ravi, Valentin Gabeur, Yuan-Ting Hu, Ronghang Hu, Chaitanya Ryali, Tengyu Ma, Haitham Khedr, Roman R{\"a}dle, Chloe Rolland, Laura Gustafson, et~al.
\newblock Sam 2: Segment anything in images and videos.
\newblock \emph{arXiv preprint arXiv:2408.00714}, 2024.

\bibitem[Xiao et~al.(2024)Xiao, Yao, Li, and Chua]{xiao2024can}
Junbin Xiao, Angela Yao, Yicong Li, and Tat-Seng Chua.
\newblock Can i trust your answer? visually grounded video question answering.
\newblock In \emph{Proceedings of the IEEE/CVF Conference on Computer Vision and Pattern Recognition}, pages 13204--13214, 2024.

\bibitem[Xu et~al.(2024)Xu, Wei, Zhong, Chen, Qi, and Wu]{xu2024exploring}
Yuanxing Xu, Yuting Wei, Shuai Zhong, Xinming Chen, Jinsheng Qi, and Bin Wu.
\newblock Exploring question guidance and answer calibration for visually grounded video question answering.
\newblock In \emph{Findings of the Association for Computational Linguistics: EMNLP 2024}, pages 3121--3133, 2024.

\end{thebibliography}
}

% WARNING: do not forget to delete the supplementary pages from your submission 
\clearpage
\setcounter{page}{1}
\appendix
\renewcommand{\thesection}{\Alph{section}}
\setcounter{section}{0}

\twocolumn[{
\centering
\Large
\textbf{\thetitle}\\
\vspace{0.3em}Supplementary Material \\
\vspace{-2mm}
\begin{center}
    \includegraphics[width=0.9\linewidth]{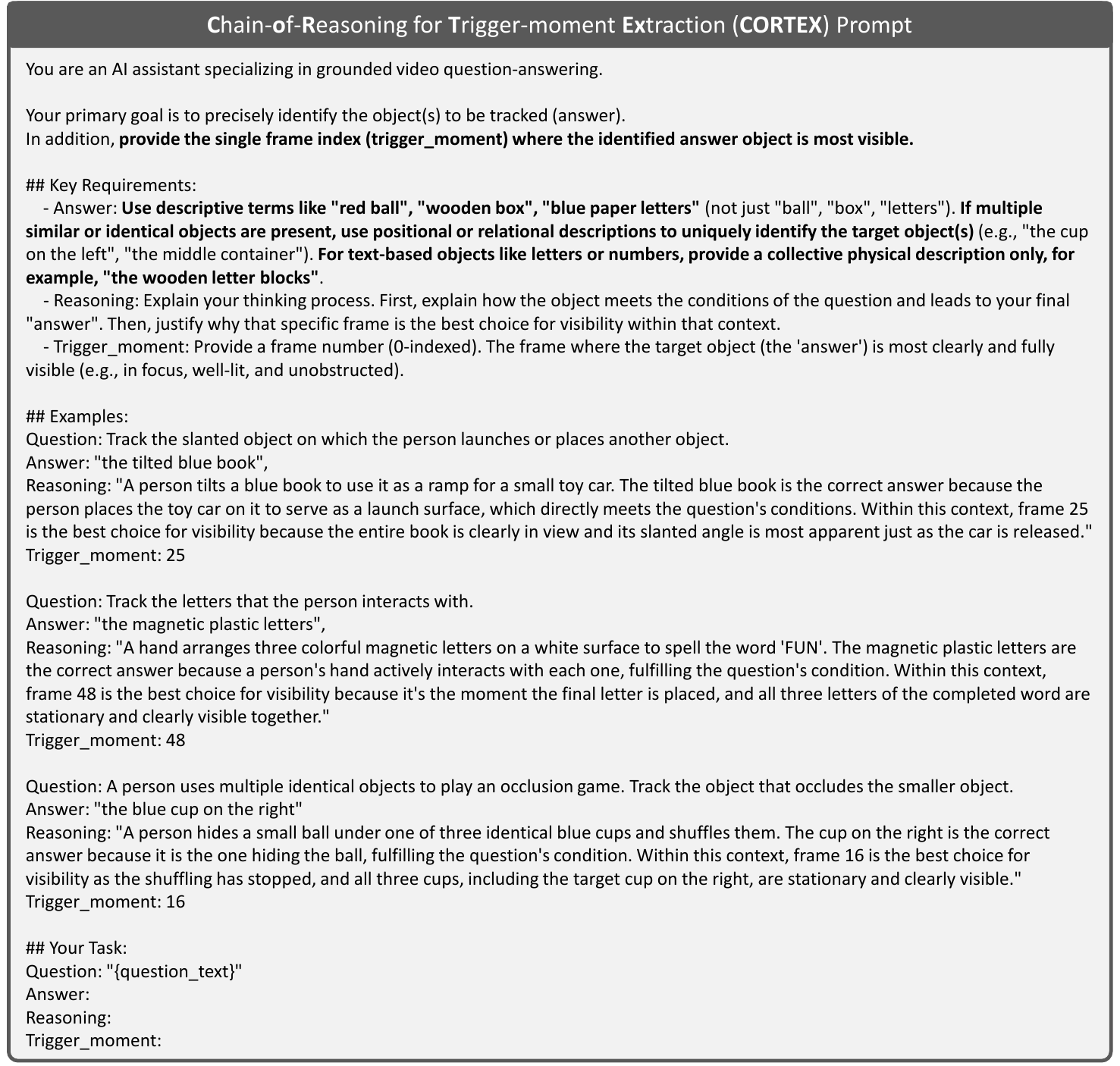}
    \vspace{-2mm}
    \captionof{figure}{\textbf{Complete template of Chain-of-Reasoning for Trigger-moment Extraction (CORTEX) prompt.} CORTEX guides the MLLM to identify target objects, explain its reasoning process, and specify the optimal temporal moment for spatial grounding.}
    \label{fig:prompt_template}
\end{center}
}]

% \clearpage

\section{Details of Prompt Design}
\label{sec:prompt_details}
\vspace{-3mm}
This section presents the complete CORTEX prompt template (Figure~\ref{fig:prompt_template}) and its underlying design principles. The prompt is built on three key principles: (1) \textit{explicit descriptive requirements} provide discriminative detail for accurate spatial grounding, (2) \textit{structured reasoning} connects video understanding to answer generation through explicit justification, and (3) \textit{visibility-focused frame selection} optimizes grounding initialization by selecting frames where target objects are clearly and fully visible.

\subsection{Prompt Structure}
The CORTEX template consists of three components: task instructions, key requirements, and few-shot examples.

\vspace{1mm}
\noindent\textbf{Task Instructions.}\hspace{2mm}This component defines the assistant's role: \textit{``You are an AI assistant specializing in grounded video question-answering...''} This frames the task as object identification with temporal localization.

\vspace{1mm}
\noindent\textbf{Key Requirements.}\hspace{2mm}This component specifies the output format for three fields:

\begin{itemize}[leftmargin=*, itemsep=1pt, parsep=0pt]
    \item \textit{Answer}: Must include visual attributes (\eg color, material). Requires positional descriptions for similar objects (\eg ``the cup on the left''). For text objects, describes the physical medium (\eg ``the magnetic plastic letters'') rather than content.
    
    \item \textit{Reasoning}: Two-step structure: (1) explain how the object satisfies the question, (2) justify why the selected frame provides optimal visibility.
    
    \item \textit{Trigger\_moment}: Single 0-indexed frame where the target is most clearly visible, considering focus, lighting, and occlusion.
\end{itemize}

\vspace{1mm}
\noindent\textbf{Few-Shot Examples.}\hspace{2mm}Three carefully selected examples demonstrate:

\begin{itemize}[leftmargin=*, itemsep=1pt, parsep=0pt]
    \item \textit{Functional Description}: ``Track the slanted object...'' → ``the tilted blue book'' (combining physical state and color).
    
    \item \textit{OCR Case}: ``Track the letters...'' → ``the magnetic plastic letters'' (describing medium, not spelling ``F-U-N'').
    
    \item \textit{Spatial Disambiguation}: ``Track the object that occludes...'' → ``the blue cup on the right'' (using position among identical objects).
\end{itemize}

\subsection{Design Rationale}

\noindent\textbf{Explicit Descriptive Requirements.}\hspace{2mm}Requiring visual attributes (\eg ``the tilted blue book'' not ``book'') provides discriminative information for accurate spatial grounding. This ensures Molmo receives sufficient detail to localize the correct object, especially when multiple similar objects are present. The requirement for positional descriptions (\eg ``the cup on the left'') further enables distinguishing between identical or highly similar objects through spatial relations.

\vspace{1mm}
\noindent\textbf{Structured Reasoning.}\hspace{2mm}The two-step reasoning requirement strengthens video understanding and QA capability. First, the model must verify the object satisfies the question's conditions, ensuring semantic correctness. Second, it must justify frame selection based on visibility, connecting high-level understanding to low-level visual grounding. This explicit reasoning reduces spurious selections and improves answer quality.

\vspace{1mm}
\noindent\textbf{Visibility-Focused Frame Selection.}\hspace{2mm}Instructing the model to select frames where objects are ``clearly and fully visible (\eg in focus, well-lit, and unobstructed)'' is crucial for successful grounding initialization. This guidance steers the model away from frames with occlusion or motion blur, selecting instead stable moments optimal for tracking initiation. The examples reinforce this: frame 25 when ``the entire book is clearly in view'', frame 48 when ``all three letters are stationary and clearly visible together'', frame 16 when ``shuffling has stopped''.

\vspace{1mm}
\noindent\textbf{OCR Handling.}\hspace{2mm}The explicit instruction to describe text objects by physical properties (\eg ``wooden letter blocks'') rather than content addresses vision-language models' weakness in character-level grounding. This redirection from semantic content to physical objects significantly improves grounding success for text-based queries.

\end{document}